\pdfoutput=1
\documentclass[11pt]{article}

\usepackage[]{EMNLP2022}

\usepackage{times}
\usepackage{latexsym}

\usepackage[T1]{fontenc}
\usepackage[utf8]{inputenc}

\usepackage{microtype}

\usepackage{booktabs}       % professional-quality tables
\usepackage{amsfonts}       % blackboard math symbols
\usepackage{colortbl}
\usepackage{pgfplots}
\usepackage{amssymb}% http://ctan.org/pkg/amssymb
\usepackage{pifont}
\usepackage{amsmath}
\usepackage{multirow}
\usepackage{url}
\usepackage{xspace}
\newcommand{\model}{PageSum\xspace}

\DeclareUnicodeCharacter{0307}{}
\newcommand{\RomanNumeralCaps}[1]{\MakeUppercase{\romannumeral #1}}
\DeclareMathOperator*{\argmax}{arg\,max}

\title{Leveraging Locality in Abstractive Text Summarization}

\author{Yixin Liu$^\textbf{1}$, Ansong Ni$^\textbf{1}$,  Linyong Nan$^\textbf{1}$, Budhaditya Deb$^\textbf{2}$,
\\ \textbf{Chenguang Zhu}$^\textbf{2}$, \textbf{Ahmed H. Awadallah}$^\textbf{2}$, \textbf{Dragomir Radev}$^\textbf{1}$\\
$^1$Yale University, $^2$Microsoft Research \\
\texttt{\{yixin.liu, ansong.ni, linyong.nan, dragomir.radev\}@yale.edu} \\
\texttt{\{Budha.Deb, chezhu, hassanam\}@microsoft.com}
}

\begin{document}
\maketitle

\begin{abstract}
Neural attention models have achieved significant improvements on many natural language processing tasks. However, the quadratic memory complexity of the self-attention module with respect to the input length hinders their applications in long text summarization. 
Instead of designing more efficient attention modules, we approach this problem by investigating if models with a \textit{restricted} context can have competitive performance compared with the memory-efficient attention models that maintain a global context by treating the input as a single sequence.
Our model is applied to individual \textit{pages} which contain parts of inputs grouped by the \textit{principle of locality} during both encoding and decoding.
We empirically investigated three kinds of locality in text summarization at different levels of granularity, ranging from sentences to documents.
Our experimental results show that our model has a better performance compared with strong baselines with efficient attention modules, and our analysis provides further insights into our locality-aware modeling strategy.\footnote{We have made our code, results, and trained models publicly available at \url{https://github.com/yixinL7/PageSum}.}

\end{abstract}

\section{Introduction}

Neural abstractive summarization~\citep{rush-etal-2015-neural, nallapati-etal-2016-abstractive} is mainly formulated as a sequence-to-sequence~\citep{Sutskever2014SequenceTS} (Seq2Seq) problem.
Neural attention models, e.g., Transformers~\citep{NIPS2017_3f5ee243}, have been widely used for such Seq2Seq tasks, allowing effective modeling of various dependencies in the input and output sequences.
However, the \textit{self-attention} module in such models introduces a quadratic memory growth with respect to the input sequence length.
Consequently, for long-text summarization datasets,\footnote{For example, the average input document length in the arXiv dataset~\citep{cohan-etal-2018-discourse} is more than 8,000 tokens.} recent works~\citep{DBLP:journals/corr/abs-2004-05150, Kitaev2020Reformer:, NEURIPS2020_c8512d14} have explored using \textit{efficient attention} to reduce the memory footprint while still maintaining the same \textbf{global context} of a full-attention model -- every input token can receive information from all the other input tokens.
However, efficient attention is just an approximation of full attention and can show lower performance compared with its counterpart~\citep{Kitaev2020Reformer:}.
To investigate an alternative memory-efficient modeling approach, we argue that models with a \textit{restricted} context, where each token only receives a subset of tokens as its context during the entire computation, can be competitive with efficient attention models if they can effectively leverage \textbf{locality} in text summarization. 

\begin{figure}[t!]
    \centering
    \includegraphics[width=1\linewidth]{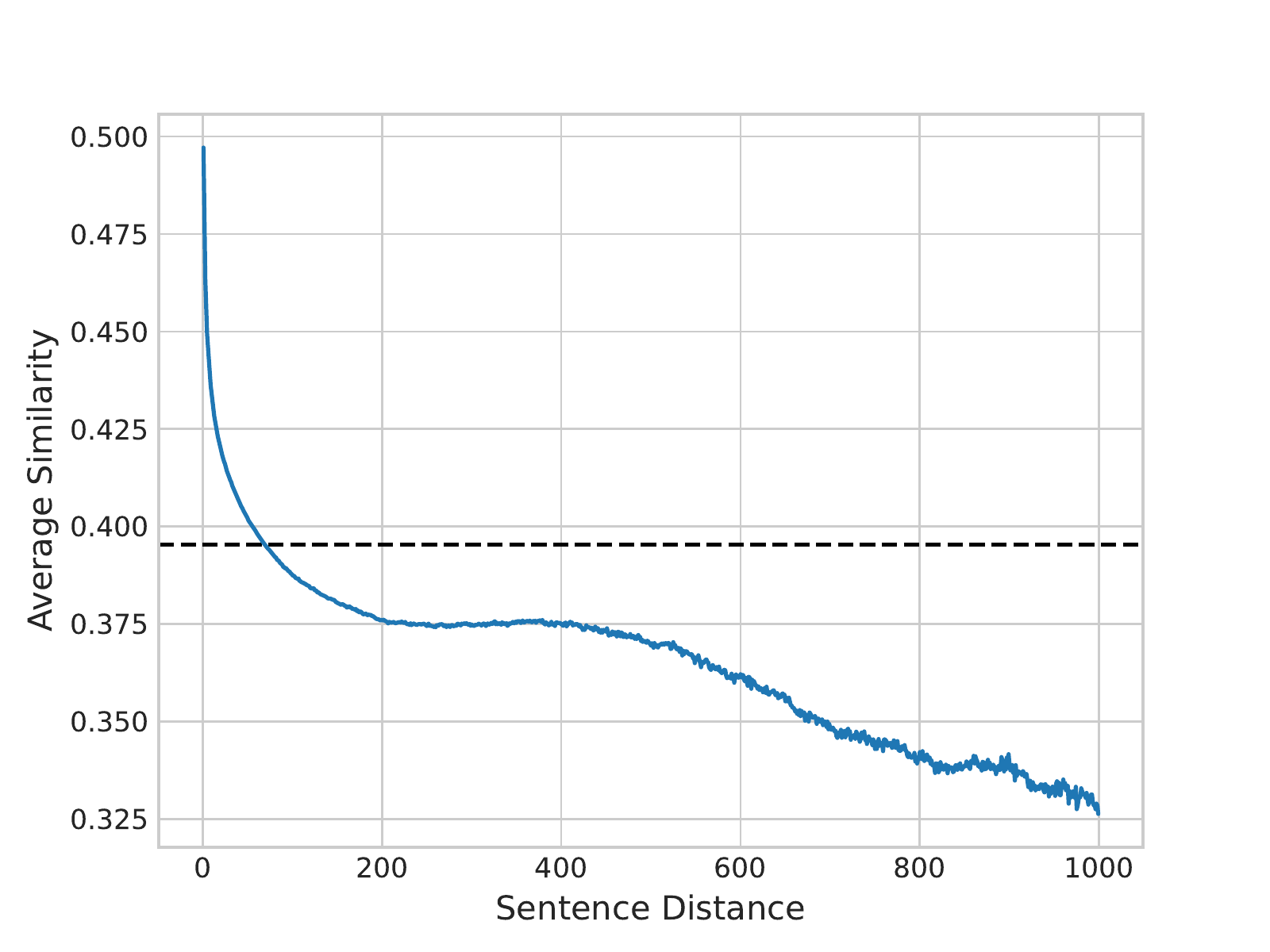}
    \caption{Intrinsic spatial locality in the arXiv dataset. The X-axis represents the distance of two sentences in source documents measured by the difference of their locations (indexes). Y-axis represents the average semantic similarity calculated by the cosine similarity between sentence embeddings, which are generated by a pre-trained sentence embedding model~\citep{gao-etal-2021-simcse}. The dash line shows the average similarity.}
    \label{fig:intro}
    \vspace{-1mm}
\end{figure}

Locality, or the principle of locality, is one of the fundamental principles of virtual memory systems~\citep{10.1145/1070838.1070856},\footnote{A formal definition of locality coined by \citet{1702696} is: ``The concept that a program favors a subset of its segments during extended intervals (phases) is called locality."} and exists in a wide range of domains~\cite{koopman2013introduction, 1208696, 10.1145/2723372.2723718}.
A classic example of locality is the \textit{spatial locality} in computer memory systems -- data units that are stored closely on the disk are likely to be accessed during a short time period by a computer process, therefore it is beneficial to read a block of data as a \textit{page} in the memory instead of reading only one data unit at a time.
Such patterns also exist in text summarization.
For example, on the arXiv dataset, we observe an \textit{intrinsic} spatial locality in source documents -- the closer in the document two sentences are, the more semantically similar they are (Fig.~\ref{fig:intro}). 
This observation supports the inductive bias of \textit{window attention}~\citep{DBLP:journals/corr/abs-2004-05150, NEURIPS2020_c8512d14}, which allows each token to interact with its neighboring tokens within the window size. 

\begin{figure}[t!]
    \centering
    \includegraphics[width=1\linewidth]{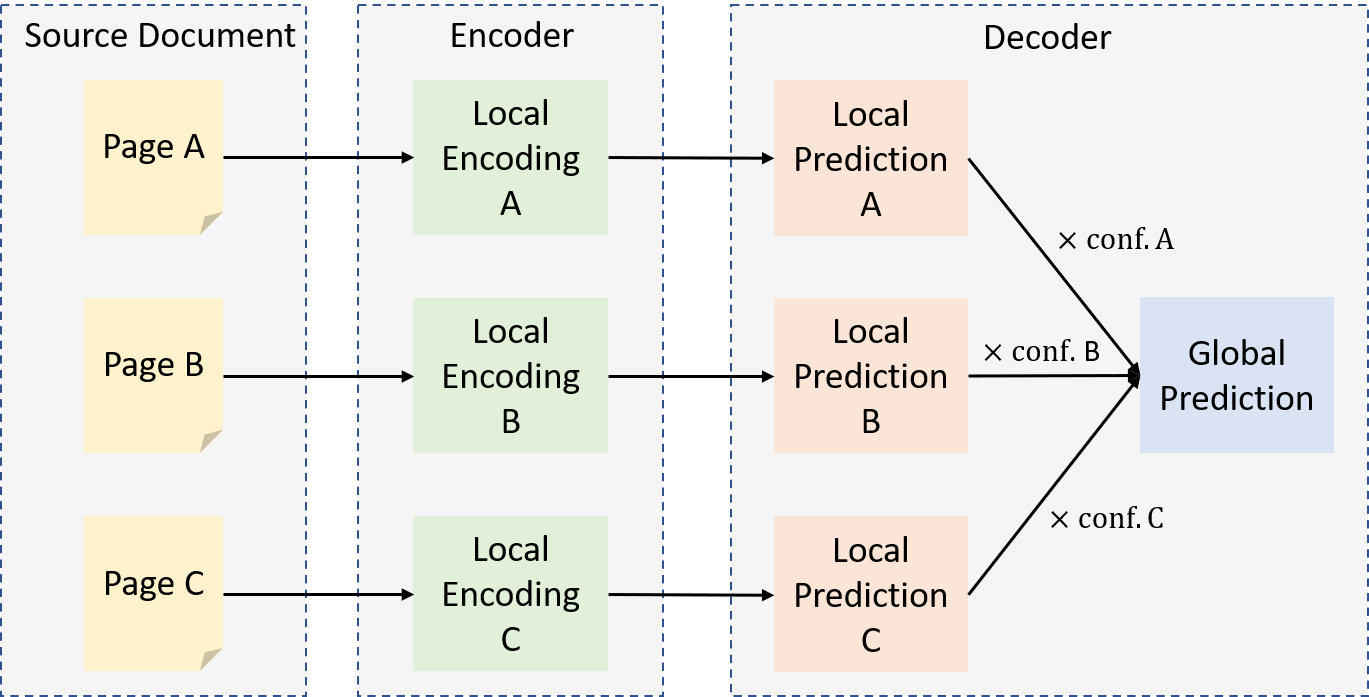}
    \caption{Model architecture. Our model views the source document as a number of \textit{non-overlapping pages}, and the final output is a weighted combination of local predictions on the individual pages.}
    \label{fig:model}
    \vspace{-3mm}
\end{figure}

We introduce a framework of leveraging locality for text summarization, which reduces the memory complexity of full-attention models while still maintains competitive performance.
Instead of viewing the input document as an entire sequence, we represent an input document as a number of \textit{pages} which are constructed according to the principle of locality (Fig.~\ref{fig:model}). 
Each of these pages is encoded independently by the encoder of our abstractive model, and the decoder makes \textit{local} predictions over each page along with \textit{local} confidence scores of its predictions, which are used to combine the local predictions into final outputs.
In this framework, tokens in different pages never directly interact with each other during encoding and decoding, which highlights the role of \textbf{locality} in text summarization.
In contrast, one of the key assumptions of efficient attention models is that all tokens in the input text should interact with each other, which is made possible because  (1) \textit{global tokens}~\citep{DBLP:journals/corr/abs-2004-05150} or \textit{overlapping} window attention maintain a \textit{global} context during encoding; (2) the encoder-decoder attention takes the source document embeddings as an entire sequence during decoding.

Using the proposed framework, we are able to investigate several types of locality in text summarization:
(1) \textit{spatial} locality or \textit{sequential} locality -- neighboring sentences are grouped into the same (non-overlapping) page;
(2) \textit{discourse} locality -- different sections in a scientific paper may cover different aspects, therefore they are viewed as different pages~\citep{cohan-etal-2018-discourse};
(3) \textit{document} locality -- for multi-document summarization, each document in a document cluster can be viewed as an individual page~\citep{jin-wan-2020-abstractive}.
Our approach also has other advantages:
(1) Our model can take the most advantage of pre-trained full-attention models (e.g. BART~\cite{lewis-etal-2020-bart}) because it preserves the same attention mechanism as the full-attention models, unlike most of the efficient attention models;
(2) It reduces the overall complexity of encoder self-attention to a \textit{linear} relationship with the input document length. 
We empirically demonstrate that our model outperforms strong baseline models built upon various efficient-attention modules on several summarization datasets.
Furthermore, we conduct detailed analyses on different modeling options for our framework, shedding lights on its broader uses.

\section{Preliminaries}

Abstractive summarization models aim to generate a shorter text sequence as the summary of an input document.
Given an input document $D$ and a reference summary $S$, the standard training algorithm of a neural abstractive summarization model $g$ adopts the cross-entropy loss, which requires the model to predict the next token of the reference summary given the input document and the prefix of the reference summary before the current token:
\begin{equation}
    \mathcal{L}_{xent} =  - \sum_{i=1}^l \log p_{g_\theta} (s_i |D, S_{<i}; \theta),
\label{eq:xent}
\end{equation}
where $\theta$ is the trainable parameters of the model $g$, $p_{g_\theta}$ is the predicted probability over the vocabulary, $l$ is the length of the summary $S$, $\{s_1, \cdots , s_i, \cdots , s_l\}$ are tokens in $S$, $S_{<i}$ denotes the partial reference sequence $\{s_0, \cdots, s_{i - 1}\}$ and $s_0$ is a pre-defined start token.

\paragraph{Encoder-Decoder Model} The encoder-decoder model formulates abstractive summarization as a Seq2Seq task,
\begin{equation}
    h_i = \mathrm{Decoder}(\mathrm{Encoder}(D) , S_{<i}),
\end{equation}
where $h_i$ is the hidden representation.
The generation probability is 
\begin{equation}
    p_{g_\theta} (\cdot|D, S_{<i}; \theta) = \mathrm{softmax}(L_{vocab}(h_i)),
\end{equation}
where $L_{vocab}$ is a linear projection layer.

\paragraph{Neural Attention and Its Limitations} Neural attention modules are essential to the success of Transformers~\citep{NIPS2017_3f5ee243} and pre-trained language models~\citep{radford2019language, lewis-etal-2020-bart, zhang2020pegasus} for language generation tasks such as machine translation or text summarization. 
Given a query matrix $Q$, a key matrix $K$, and a value matrix $V$, the output of the dot-product attention is: 
\begin{equation}
\small
\label{eq:att}
  \mathrm{Attention}(Q, K, V) = \mathrm{softmax}(QK^{T})V.
\end{equation}
To compute Eq.~\ref{eq:att} in a parallel manner, it requires $\mathcal{O}(l_{Q} \cdot l_{K})$ memory space to store the intermediate result of $QK^{T}$ where $l_{Q}$ and $l_{K}$ are the length of $Q$ and $K$ respectively. 
This becomes a bottleneck of the \textit{self-attention} module for long input documents, where $Q$, $K$, $V$ come from the same input $D$, and the space complexity becomes $\mathcal{O}({l_{D}}^2)$, where $l_D$ is the length of the input document and can be very large (e.g. more than 10,000 tokens).

\section{Locality-aware Abstractive Text Summarization}

To avoid the quadratic growth of memory with respect to the length of the input, we introduce a different view for modeling the input text. 
Specifically, instead of viewing the input document as an entire text sequence, we view it as a series of \textit{non-overlapping pages} with a fixed maximum length:
\begin{equation}
    D := \{P_1, \cdots, P_i, \cdots, P_n\},
\end{equation}
where $P_i$ is the $i$-th page and $n$ is the number of pages.
We hypothesize that with the \textbf{principle of locality}, the abstractive summarizer can make local predictions about the output summary based on \textit{individual} pages without having each input token interact with the entire input document:
\begin{equation}
\label{eq:page}
    h_i^{(j)} = \mathrm{Decoder}(\mathrm{Encoder}(P_j) , S_{<i}),
\end{equation}
where $h_i^{(j)}$ is the \textit{local} hidden state of the $i$-th token of the summary given the $j$-th page.
Apart from the hidden state, we also require the decoder to predict a confidence score of its local prediction:
\begin{equation}
\label{eq:conf}
    c_{ij} = L_{conf}( h_i^{(j)} ),
\end{equation}
where $L_{conf}$ is a linear layer projecting the hidden state $h_i^{(j)}$ to a scalar. 
The confidence scores are normalized:
\begin{equation}
    \hat{c}_{ij} = \frac{\exp(c_{ij})}{\sum_{k=1}^n \exp(c_{ik})},
\end{equation}
and used to combine the local hidden states for predicting the final output:
\begin{equation}
\label{eq:global}
\small
    p_{g_\theta} (\cdot|D, S_{<i}; \theta) = \mathrm{softmax}( L_{vocab}(\sum_{j=1}^n \hat {c}_{ij} \cdot h_i^{(j)})).
\end{equation}

\paragraph{Fine-tuning from Pre-trained Models}
Our model can be directly initialized from a pre-trained language model (e.g. BART~\cite{lewis-etal-2020-bart}) except for an additional linear layer $L_{cong}$ (Eq.~\ref{eq:conf}).
The cross-entropy loss (Eq.~\ref{eq:xent}) with label smoothing~\cite{7780677} is used for training.

\paragraph{Space Complexity} Our model has a linear space complexity with respect to the length of input documents.
Specifically, given a pre-defined maximum page length $L_{page}$, a document of which the length is $l_{D}$ will be split into at most $\lceil \frac{l_{D}}{L_{page}} \rceil$ pages.
The space complexity of the encoder self-attention for one page is $\mathcal{O}(L_{page}^2)$, and the complexity for all pages is
\begin{equation}
    \mathcal{O}(L_{page}^2 \cdot \lceil \frac{l_{D}}{L_{page}} \rceil ) = \mathcal{O}(L_{page}l_{D}).
\end{equation}
When $l_{D}  \gg  L_{page}$, the complexity is $\mathcal{O}(l_{D})$.\footnote{In practice, the page size $L_{page}$ can be large (e.g. 512 tokens). However, we note that sparse attention models can also use window attention with large sizes (e.g. Longformer~\citep{DBLP:journals/corr/abs-2004-05150} uses either 512 or 1024 tokens).}
\paragraph{Locality in Abstractive Summarization}

We mainly explore three types of locality for abstractive summarization, which provide the principles of splitting an input document or document cluster (in the case of multi-document summarization) into different \textit{pages}.

\noindent (1) \textbf{Spatial Locality}: in the most direct form, an input document can be sequentially split into different pages. The underlying intuition is that neighboring sentences are likely to focus on the same topic.
Under this setting, each document will be equally split into $n_{p}$ pages, which is a pre-defined number.

\noindent (2) \textbf{Discourse Locality}: long documents usually have a hierarchical discourse structure, and discourse units at the same level have different focus. 
For example, a scientific paper usually has multiple sections with different purposes (e.g. introduction, related work, etc.), and this discourse structure can be a useful inductive bias~\cite{cohan-etal-2018-discourse}.
Under this setting, each discourse unit (e.g. a section in a scientific paper) is viewed as a page.

\noindent (3) \textbf{Document Locality}: for multi-document summarization, we can view each single document in the document cluster as a page.
Previous work~\citep{jin-wan-2020-abstractive} has shown that multi-document summarization can benefit from single-document summarization models by first summarizing each document then combining the predictions.

\section{Related Work}

\subsection{Efficient Attention Models}

Efficient attention models aim to reduce the memory complexity of full attention models, of which the most important and commonly used building blocks are window attention~\citep{DBLP:journals/corr/abs-2004-05150, NEURIPS2020_c8512d14} and low-rank approximation~\citep{j.2018generating, Wang2020LinformerSW, peng2021random, choromanski2021rethinking}.

Window attention means that each token can only receive information from its neighboring tokens that are located in the same window.
However, multi-layer models with \textbf{overlapping} window attention~\citep{DBLP:journals/corr/abs-2004-05150, NEURIPS2020_c8512d14, manakul-gales-2021-long, Guo2021LongT5ET} can still maintain a \textit{global} context.
On the other hand, \textbf{non-overlapping} window attention (local attention) with \textit{fixed} windows~\citep{j.2018generating, Zhao2020SEALSE,Pietruszka2020SparsifyingTM} has a \textit{restricted} context since tokens in different windows cannot interact with each other.
Instead of using fixed windows throughout the model, using window attention with \textbf{learnable} patterns~\citep{Kitaev2020Reformer:, pmlr-v119-tay20a, huang-etal-2021-efficient} offer more flexibility because windows can be dynamically constructed at different layers of the model, which allows a larger context.
Headwise sparse attention~\citep{qiu-etal-2020-blockwise, huang-etal-2021-efficient} is another method of reducing memory usage while preserving global context.

Compared to these methods, our model has a distinct feature in that we maintain a \textbf{local} context of the input tokens at both the encoding and decoding stages.
\citet{Zhao2020SEALSE} proposed a similar block-wise encoder-decoder attention module which only uses a subset of input tokens (blocks) at each decoding stage. 
However, our method differs from theirs in that our model dynamically combines the local predictions based on all the individual \textit{pages} into the final output (Eq.~\ref{eq:global}). 

\subsection{Hierarchical Summarization Models}

Hierarchical attention~\citep{yang-etal-2016-hierarchical} models aim to utilize the inherent structure of documents as a source of inductive bias.
For text summarization, \citet{ling-rush-2017-coarse} proposes a coarse-to-fine structure consisting of word-level and chunk-level attention. 
\citet{cohan-etal-2018-discourse, xu-etal-2020-discourse, dong-etal-2021-discourse} introduce discourse-aware attention at the level of document sections or elementary discourse units.
Related work~\citep{ xiao-carenini-2019-extractive,xu-etal-2020-unsupervised, DBLP:journals/corr/abs-2104-07545, ruan-etal-2022-histruct} use a similar structure that computes both token-level and sentence-level attention.
\citet{cao-wang-2022-hibrids} introduces learnable hierarchical biases into the attention module.

Hierarchical models have also been widely used for multi-document summarization. 
Hierarchical attention can focus on the sentence level~\citep{fabbri-etal-2019-multi}, paragraph level~\citep{liu-lapata-2019-hierarchical}, and document level~\citep{zhang-etal-2018-adapting, jin-wan-2020-abstractive,jin-etal-2020-multi}.
\citet{Ernst2021APC} porposed a proposition-level clustering algorithm, which generates summaries from each of the proposition clusters extracted from source documents. 

The multi-stage method of text summarization~\citep{chen-bansal-2018-fast, xu-durrett-2019-neural, pilault-etal-2020-extractive} also has a hierarchical structure. 
In particular, \citet{zhang-etal-2022-summn} first generates a coarse summary for each part of the input document, then further summarizes the generated summaries.
\citet{mao-etal-2022-dyle} first extracts sentences from the source documents, and generates the summary based on the selected sentences.

Our method introduces \textbf{pages} as a new, unified abstraction for hierarchical models which can be instantiated as sentence clusters, scientific paper sections, and entire documents in a document cluster.
Furthermore, unlike previous work, our model emphasizes the role of locality by preventing explicit interactions among different units (pages) at the higher levels of the hierarchy.

\begin{table*}[t]
  \centering
  \small
    \begin{tabular}{@{\extracolsep{1pt}}lccccccccc}
    \toprule
    \multirow{2}{*}{\textbf{System}} & \multicolumn{3}{c}{\textbf{arXiv}} & \multicolumn{3}{c}{\textbf{PubMed}} & \multicolumn{3}{c}{\textbf{GovReport}}  \\
    \cmidrule{2-4} \cmidrule{5-7} \cmidrule{8-10}
    & \textbf{R-1} & \textbf{R-2} & \textbf{R-L} & \textbf{R-1} & \textbf{R-2} & \textbf{R-L} & \textbf{R-1} & \textbf{R-2} & \textbf{R-L}\\
    \midrule
    LED* (4096) & 44.40 & 17.94 & 39.76 & - & - & - & - & - & - \\
    LED* (16384) & 46.63 & 19.62 & 41.83 & - & - & - & - & - & - \\
    $\textrm{LED}^\ddag$ (16384) & 48.10 & 19.78 & 43.08 & 46.93 & 19.88 & 42.73 & 59.42 & 26.53 & 56.63  \\
    HEPOS* (7168) & 48.24 & 20.26 & 41.78 & 48.12 & 21.06 & 42.72 & 55.00 & 21.13 & 51.67 \\
    HEPOS* (10240) & 47.87 & 20.00 & 41.50 & 47.93 & 20.74 & 42.58 & 56.86 & 22.62 & 53.82 \\
    PRIMERA* (4096) & 47.60 & 20.80 & 42.60 & - & - & - & - & - & -
    \\
    $\textrm{PRIMERA}^\ddag$ (4096) & 47.65 & 20.76 & 43.19 & - & - & - & - & - & - \\
    HAT-BART*  (3072) & 46.68 & 19.07 & 42.17 & \textbf{48.36} & \textbf{21.43} & 37.00 & - & - & - \\
    \midrule
    \model (7168)  & $\textbf{49.72}^{\dag}$ & $\textbf{21.06}^{\dag}$ & $\textbf{44.69}^{\dag}$ & $48.24^{\dag}$ & $21.06^{\dag}$ & $44.26^{\dag}$ & 59.05 & 26.37 & 56.22 \\
    \model (20480)  & - & - & - & - & - & - & $59.91^{\dag}$ & $\textbf{27.20}^{\dag}$ & $57.07^{\dag}$ \\
    $\textrm{\model}^\star$ (7168/20480) & $49.60^{\dag}$ & $20.98^{\dag}$ & $\textbf{44.69}^{\dag}$ & $48.73^{\dag}$ & $21.33^{\dag}$ & $\textbf{44.67}^{\dag}$ & $\textbf{60.04}^{\dag}$ & $27.17^{\dag}$ & $\textbf{57.21}^{\dag}$ \\
    \bottomrule
    \end{tabular}%
  \caption{System performance comparison for \textit{spatial locality}. R-1/2/L are the ROUGE-1/2/L $\mathrm{F_1}$ scores respectively. The numbers in parentheses indicate the maximum input length (tokens).  *: results reported in the original papers. \ddag: results from our own evaluation script (and own checkpoints). $\dag$: significantly better than $\textrm{LED}^\ddag$ ($p < 0.01$). \model denotes the model fine-tuned from a BART checkpoint pre-trained on the CNN/DailyMail dataset, while $\textrm{\model}^\star$ is its counterpart without the CNN/DailyMail pre-training. For $\textrm{\model}^\star$, the maximum token number is 7168 on \texttt{arXiv} and \texttt{PubMed}, and 20480 on \texttt{GovReport}. } 
  \label{tab:exp-1}%
  \vspace{-3mm}
\end{table*}%

\section{Experiments}

\subsection{Experimental Settings}
\label{subsec:setting}

\paragraph{Datasets} We use four datasets (Tab.~\ref{tab:data}) in our experiments. 

\noindent \texttt{arXiv} and \texttt{PubMed} are two scientific paper summarization datasets introduced by \citet{cohan-etal-2018-discourse}.\footnote{\url{https://github.com/armancohan/long-summarization}}
The abstracts of the papers are used as the summaries of the main content of those papers.

\noindent \texttt{GovReport}\footnote{\url{https://github.com/luyang-huang96/LongDocSum}}~\cite{huang-etal-2021-efficient} is a long document summarization dataset based on reports published by the U.S. Government Accountability Office and Congressional Research Service.

\noindent \texttt{MultiNews}\footnote{\url{https://github.com/Alex-Fabbri/Multi-News}}~\cite{fabbri-etal-2019-multi} is a multi-document summarization dataset, with news articles and summaries collected from newser.com.

\paragraph{Baselines} 
We use the following top-performing models as baselines for comparison.

\noindent (1) \textbf{LED} (Longformer Encoder-Decoder)~\citep{DBLP:journals/corr/abs-2004-05150} is an encoder-decoder model with a sparse encoder self-attention module.

\noindent (2) \textbf{HEPOS}~\cite{huang-etal-2021-efficient} combines both efficient encoder self-attention and encoder-decoder attention in its encoder-decoder architecture.

\noindent (3) \textbf{PRIMERA}~\citep{xiao-etal-2022-primera} shares the same architecture as \textbf{LED}, but has task-specific pre-training for multi-document summarization.

\noindent (4) \textbf{HAT-BART}~\citep{DBLP:journals/corr/abs-2104-07545} is built upon BART~\citep{lewis-etal-2020-bart} while it has additional hierarchical layers for sentence-level interactions. 
It uses \textit{full} attention and not sparse attention.

\paragraph{Implementation Details}

We use BART\footnote{It contains around 400M parameters.} as the backbone of our model, except for the linear layer computing the confidence scores (Eq.~\ref{eq:conf}). 
We initialize the model from either a checkpoint pre-trained on CNN/DailyMail dataset~\citep{10.5555/2969239.2969428, nallapati-etal-2016-abstractive}, or its counterpart without the CNN/Dailymail pre-training. 
We select the model checkpoints based on their performance on the validation set, using cross-entropy loss (Eq.~\ref{eq:xent}).
We use ROUGE~\citep{lin-2004-rouge} as the automatic evaluation metric for performance comparison.
More specifically, we report the F1 scores of ROUGE-1/2/L in our experiments.  

We name our model as \textbf{\model} for the following experiments.

\subsection{Exp-\RomanNumeralCaps{1}: Spatial Locality}
\label{subsec:exp-1}

We first investigate the case of \textit{spatial locality}, where the sentences in the source document are sequentially split into different \textit{pages} with the same number of sentences.
The maximum number of tokens for one page is 1,024.

We report the model performance\footnote{For a fair comparison, we used public-available checkpoints of LED from Hugging Face's Transformers~\citep{wolf-etal-2020-transformers} on \texttt{arXiv} (`allenai/led-large-16384-arxiv') and \texttt{PubMed} (`patrickvonplaten/led-large-16384-pubmed') to generate the summaries and used our own evaluation script. The performance difference between the original result and the ours is likely because the original implementation uses window-attention with 512 tokens while HF uses 1,024 tokens.} in Tab.~\ref{tab:exp-1} on the \texttt{arXiv}, \texttt{PubMed}, \texttt{GovReport} datasets.
We make the following observations.
(1) \textbf{\model} achieves better ROUGE scores on all three long text summarization datasets compared with the baselines that leverage efficient attention modules.
(2) On \texttt{Pubmed}, \textbf{HAT-BART} achieves slightly better performance than \textbf{\model}, likely because HAT-BART uses \textit{full} attention instead of \textit{efficient} attention.
(3) On \texttt{GovReport}, increasing the maximum input length helps to improve \textbf{\model}'s performance.

\subsection{Exp-\RomanNumeralCaps{2}: Discourse Locality}
\label{subsec:exp-2}
\begin{table}[t!]
\centering
\small
\begin{tabular}{lccc}
\toprule
\textbf{System} & \textbf{R-1} & \textbf{R-2} & \textbf{R-L}  \\
\midrule
    \model-Spatial (7168)  & 49.72 & 21.06 & 44.69 \\
    \model-Discourse (8192)  & \textbf{49.84} & $\textbf{21.19}^{\dag}$ & $\textbf{44.89}^\dag$  \\
\bottomrule
\end{tabular}
\caption{System performance comparison for \textit{discourse locality} on \texttt{arXiv}. R-1/2/L are the ROUGE-1/2/L $\mathrm{F_1}$ scores respectively. The numbers in the parentheses indicate the maximum input length. \textbf{\model-Spatial} is with spatial locality. \textbf{\model-Discourse} is with discourse locality. $\dag$: significantly better ($p < 0.05$).}
\label{tab:exp-2} 
\vspace{-2mm}
\end{table}

\begin{table}[t]
\centering
\begin{tabular}{cccc}
\toprule
\textbf{reference} & \textbf{random} & \textbf{spatial} & \textbf{discourse}  \\
\midrule
 0.9800 & 0.9543 & 0.9734 & 0.9798  \\
\bottomrule
\end{tabular}
\caption{Semantic coherence (Eq.~\ref{eq:coherent}) of summaries on \texttt{arXiv}. 
\textbf{reference} is the reference summary. \textbf{random} is an oracle which randomly shuffles reference summary sentences. \textbf{spatial} is \model with spatial locality while \textbf{discourse} is with discourse locality.\textbf{discourse} has significantly higher ($p < 0.01$) coherence than \textbf{spatial}.}
\label{tab:coherent} 
\vspace{-5mm}
\end{table}

We use the \texttt{arXiv} dataset to explore another locality principle -- \textit{discourse locality}.
Specifically, we view each section of the input document as an individual \textit{page}.
The maximum number of tokens for one \textit{page} is still 1,024, however, here we allow each example to have a different number of pages because documents can have different numbers of sections.
For each page, we concatenate the name of the section and the content together as the input.

The results in Tab.~\ref{tab:exp-2} show that \model with \textit{discourse locality} achieves higher ROUGE scores than \model with \textit{spatial locality}.
In addition, we note that with discourse locality, \model can also generate more coherent summaries.
Specifically, following \citet{bommasani-cardie-2020-intrinsic}, we evaluate the \textit{semantic coherence} of the generated summaries using the next sentence prediction task~\citep{devlin-etal-2019-bert} with a pre-trained BERT model\footnote{We use the checkpoint (`bert-large-uncased') from HuggingFace Transformers~\citep{wolf-etal-2020-transformers}.} to predict the probability ($p_{\scriptscriptstyle \textrm{BERT}}$) of one sentence $S^{(i-1)}$ in the summary $S$ being followed by the next sentence $S^{(i)}$:
\vspace{-2mm}
\begin{equation}
\vspace{-2mm}
\label{eq:coherent}
    \mathrm{SC}(S) = \frac{\sum_{i=2}^{N_{S}} p_{\scriptscriptstyle \textrm{BERT}}(S^{(i)}|S^{(i-1)})}{N_{S} - 1},
\end{equation}
where $N_{S}$ is the number of sentences in the summary.
Tab.~\ref{tab:coherent} shows the average semantic coherence of summaries.
The summaries generated by \model with discourse locality have higher semantic coherence, suggesting that grouping the sentences based on discourse structures helps to generate more well-structured summaries. 

\subsection{Exp-\RomanNumeralCaps{3}: Document Locality}

\begin{table}[t!]
\centering
\small
\begin{tabular}{lccc}
\toprule
\textbf{System} & \textbf{R-1} & \textbf{R-2} & \textbf{R-L}  \\
\midrule
    PRIMERA*  & 49.90 & 21.10 & 25.90 \\
    $\textrm{PRIMERA}^\ddag$ & 50.29 & 21.20 & 46.23 \\
    BART-Long-Graph*  & 49.24 & 18.99 & 23.97 \\
    \midrule
    \model-Spatial & 49.03 & 19.10 & 44.73 \\
    \model-Document  & $\textbf{51.17}^\dag$ & $\textbf{21.39}^\dag$ & $\textbf{46.88}^\dag$\\
\bottomrule
\end{tabular}
\caption{System performance comparison for \textit{document locality} on \texttt{MultiNews}. R-1/2/L are the ROUGE-1/2/L $\mathrm{F_1}$ scores respectively.  \textbf{\model-Spatial} is \model with spatial locality. \textbf{\model-Document} is with document locality. *: results reported in the original papers. \ddag: results from our own evaluation script. $\dag$: significantly better than $\textrm{PRIMERA}^\ddag$ ($p < 0.05$).}
\label{tab:exp-3} 
\vspace{-5mm}
\end{table}

For multi-document summarization, we evaluate \model with \textit{document locality} on \texttt{MultiNews}, where we view each document in the document cluster as a page. 
The other experiment setting is the same as in \S\ref{subsec:exp-2}.
In addition to the baseline systems in \S\ref{subsec:setting}, we add another model BART-Long-Graph~\citep{pasunuru-etal-2021-efficiently} for comparison, which is specifically designed for multi-document summarization and achieves top performance on \texttt{MultiNews}.
The results are shown in Tab.~\ref{tab:exp-3}.\footnote{
We reported the performance fine-tuned from the BART model pre-trained on CNN/DailyMail dataset while we found the model without this pre-training having similar performance.
We notice a large difference between ROUGE-L scores reported by the original paper and as calculated using our evaluation script for PRIMERA. This may be due to different versions of ROUGE-L.}
\model achieves strong performance in this setting, outperforming the previous state-of-the-art models.
We also note that \model with \textit{document locality} achieves much better performance than its counterpart with \textit{spatial locality}, suggesting the importance of choosing the suitable locality for a specific task.

\subsection{Analysis}
We analyze several important aspects of our method to gain further insights.

\begin{table}[t!]
\centering
\small
\begin{tabular}{ccccc}
\toprule
\textbf{Page Size} & \textbf{\#Pages} & \textbf{R-1} & \textbf{R-2} & \textbf{R-L}  \\
\midrule
128 & 32 & 47.67 & 18.76 & 42.82\\
256 & 16 & 48.29 & 19.32 & 43.38\\
512 & 8 & \textbf{48.82} & 19.80 & \textbf{43.85}\\
1024 & 4 &  48.66 & \textbf{19.90} & 43.74\\
\bottomrule
\end{tabular}
\caption{Performance comparison of different \textit{page sizes} on \texttt{arXiv}. \textbf{Page Size} denotes the number of tokens in one page. 
\textbf{\#Pages} denotes the number of pages.
R-1/2/L are the ROUGE-1/2/L $\mathrm{F_1}$ scores respectively.}
\label{tab:page-size} 
\end{table}

\paragraph{Page Size} To investigate how the maximum length of a \textit{page} affects the model performance, we conduct experiments with different page sizes on \texttt{arXiv}.
For a fair comparison, we first truncate each document in \texttt{arXiv} to 4,096 tokens, then split the document into different pages based on the page size.
The results are shown in Tab.~\ref{tab:page-size}. 
We observe that increasing the page size generally helps to improve model performance.
However, model performance stops increasing after the page size reaches 512 tokens.

\begin{table}[t!]
\centering
\small
\begin{tabular}{lccc}
\toprule
\textbf{System} & \textbf{R-1} & \textbf{R-2} & \textbf{R-L}  \\
\midrule
\multicolumn{4}{c}{arXiv} \\
\midrule
    Global-Decoding & 48.57  & \textbf{19.92}  &  43.71 \\
    \model-Spatial & \textbf{48.66} & 19.90 & \textbf{43.74} \\
    \midrule
\multicolumn{4}{c}{MultiNews} \\
\midrule
    Global-Decoding & 48.75 & 19.03 & 44.48 \\
    \model-Document  & \textbf{51.17} & \textbf{21.39} & \textbf{46.88}  \\
\bottomrule
\end{tabular}
\caption{Comparison of page-wise decoding and global decoding on \texttt{arXiv} and \texttt{MultiNews}. R-1/2/L are the ROUGE-1/2/L $\mathrm{F_1}$ scores respectively.}
\label{tab:page-wise}
\vspace{-5mm}
\end{table}

\paragraph{Page-wise v.s. Global Decoding}

Both the encoder and decoder in \model are designed to follow the principle of locality. 
Specifically, the decoder in \model first makes \textit{local} predictions based on each encoded page (Eq.~\ref{eq:page}), which are later combined into final predictions.
An alternative approach is to directly make \textit{global} predictions based on the entire input document -- the encoded pages are concatenated as a single sequence, which serves as the input to the decoder. 
We compare this option with our modeling strategy in Tab.~\ref{tab:page-wise}.\footnote{On \texttt{arXiv}, we compare the models with this setting: 4 pages, 1,024 tokens for each page.}
The results show that on \texttt{arXiv}, page-wise decoding with \textit{spatial locality} has a similar performance compared with global decoding.
On the other hand, \textit{document locality} on \texttt{MultiNews} is proven to be a very useful inductive bias because \model with document locality has a large improvement over the model with global decoding.

\begin{figure}[t!]
    \centering
    \includegraphics[width=0.9\linewidth]{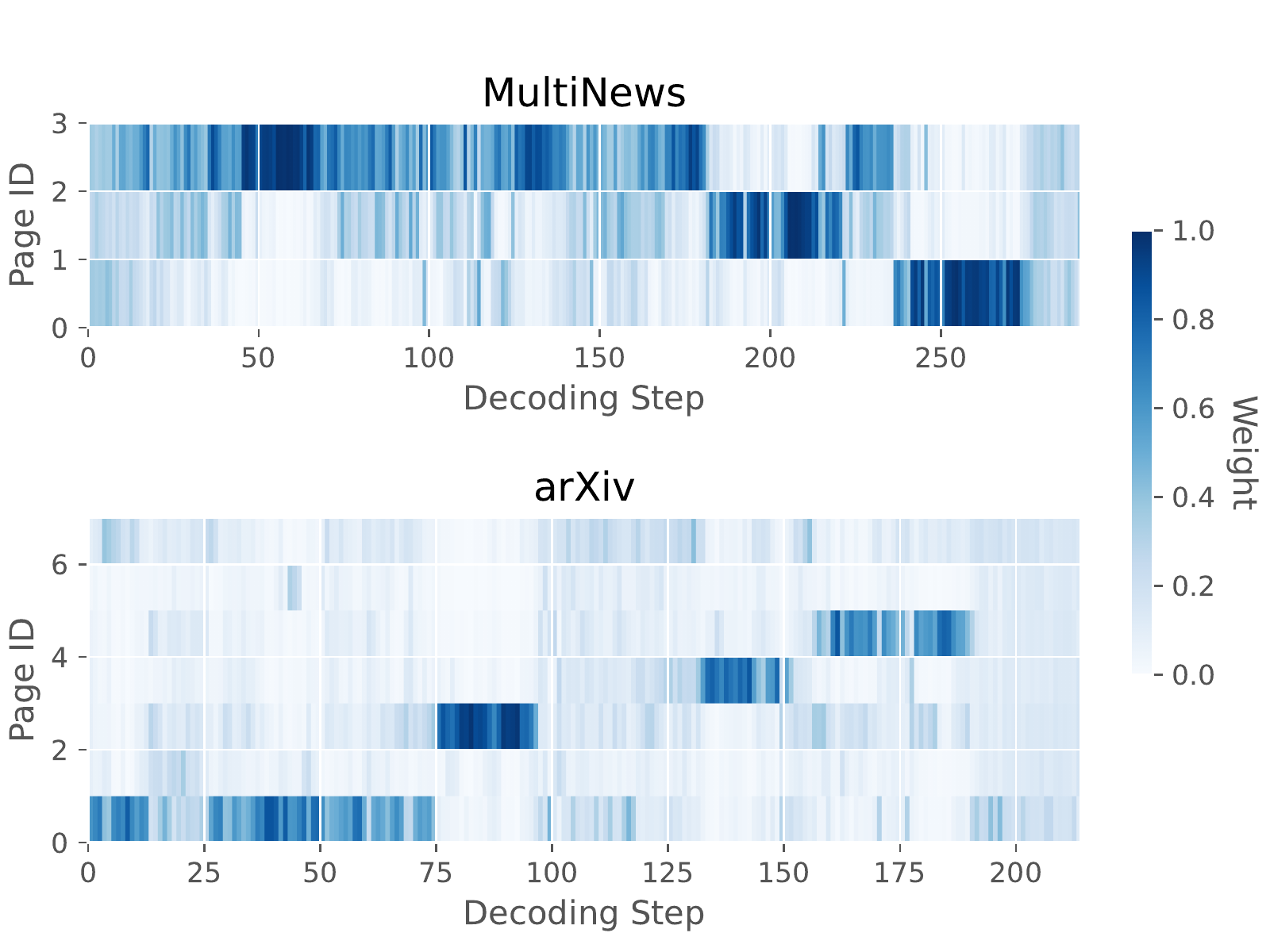}
    \caption{Visualization of importance scores of different pages at each decoding step on \texttt{MultiNews} and \texttt{arXiv}. Darker colors represent greater importance.}
    \label{fig:att}
\end{figure}

\paragraph{Visualizing Locality}
The confidence scores calculated by \model's decoder (Eq.~\ref{eq:conf}) can be interpreted as the importance scores of different \textit{pages} at each decoding step.   
That is, a page associated with a higher score will contribute more to the decision at the current step.
Fig.~\ref{fig:att} depicts how the importance scores changed during the decoding of the \textit{reference summaries} on \texttt{MultiNews} and \texttt{arXiv} using two examples.
We observe two phenomena: (1) \textit{space locality} -- at each decoding step only a subset of pages are making large contributions to the current prediction; (2) \textit{time locality} -- \model's decoder tends to focus on the similar subset of pages at neighboring decoding steps. 

\subsection{Human Evaluation for Coherence}
\label{subsec:human}

\begin{table*}[t]
    \scriptsize
    \centering
    \extrarowheight=\aboverulesep
    \addtolength{\extrarowheight}{\belowrulesep}
    \aboverulesep=0pt
    \belowrulesep=0pt
    \begin{tabular}{@{} p{0.1\linewidth} p{0.6\linewidth} p{0.2\linewidth}}
\toprule
\textbf{Error Type} & \textbf{Example} & \textbf{Explanation} \\
\midrule
\textbf{RefE} & \cellcolor{gray!25} The Part D program, administered by the Centers for Medicare \& Medicaid Services (CMS), pays Part D plan sponsors to provide drug coverage, and plan sponsors may charge beneficiaries monthly premiums in exchange for coverage. Plan sponsors and \textbf{PBMs} negotiate reimbursement rates for the drugs provided to beneficiaries. ... Seventy-four percent of the drug benefits management services provided under 624 Part D plans sponsors' contracts were performed by a \textbf{pharmacy benefit manager (PBM)} alone or in conjunction with a plan sponsor in 2016.
& The word, \textbf{PBM}, is an abbreviation for \textit{pharmacy benefit manager}, which is mentioned without first introducing the full name. \\
\textbf{TopicE} & \cellcolor{gray!10} … The President may implement the recommendations suggested in the Commerce report, take other actions, or decide to take no action. After making a decision, \textbf{the President has 15 days to implement the action and 30 days to submit a written statement to Congress explaining the action or inaction; he must also publish his findings in the Federal Register.} \textit{While there is no specific definition of national security in the statute}, it states that the investigation must consider certain factors, such as domestic production needed for projected national defense requirements; domestic capacity; …
& The topic abruptly changes from \textbf{the President and recommendations} to \textit{the specific definition of national security}.\\
\textbf{InconE} & \cellcolor{gray!25} ... To do this work, GAO selected \textit{seven} states \textbf{Arizona, Florida, Kansas, New Jersey, Pennsylvania, Tennessee, New York, Virginia, and Pennsylvania} based on factors such as population size, Medicaid enrollment, and geographic location and interviewed CMS officials. ... & There are \textbf{nine} states mentioned instead of \textit{seven}.
\\
\textbf{RepE} & \cellcolor{gray!10} … The high productivity helped the operation come in under budget by \$118 million \textbf{a 36 percent reduction} while the operation's cost was \$185 million, \textbf{36 percent below} the anticipated cost. …
& The \textbf{36 percent reduction} are mentioned twice in one sentence.\\
\bottomrule
\end{tabular}
\caption{Examples of different coherence errors on \texttt{GovReport} dataset.  \textbf{RefE}: Missing Information/Reference about an Event/Object. \textbf{TopicE}:  Abrupt Transition from the Previous Topic. \textbf{InconE}: Inconsistent, Conflicting Information. \textbf{RepE}: Repetition.}
\label{tab:coherence-example}
\vspace{-3mm}
\end{table*}

Summary coherence is a critical aspect of the summary quality, especially when the summaries are very long.
\citet{10.1162/tacl_a_00373} shows that automatic metrics have a low correlation with human evaluation results w.r.t. summary coherence, while \citet{Goyal2022SNaCCE} demonstrates that recent state-of-the-art summarization models can still make many coherence errors on long text summarization datasets.
Therefore, we conduct human evaluation for the coherence of system-generated summaries on \texttt{GovReport}\footnote{We choose \texttt{GovReport} dataset because it has the longest summaries.} dataset to investigate this important aspect.

Following \citet{Goyal2022SNaCCE}, we use a fine-grained human evaluation protocol which requires the annotators to identify different types of \textit{span-level} coherence errors in the summaries. 
We adopted the taxonomy of coherence errors proposed by \citet{Goyal2022SNaCCE} and modified it for \texttt{GovReport}, which results in four types of coherence errors (the definitions are taken and modified from the definitions in \citet{Goyal2022SNaCCE}):

(1) Missing Information/Reference about an Event/Object (\textbf{RefE}). 
These refer to coherence errors where an event or object is mentioned the first time without the proper context or introduction.
On \texttt{GovReport}, a common error is referring an entity using its abbreviation without introducing the entity and its whole name before.  

(2) Abrupt Transition from the Previous Topic (\textbf{TopicE}). 
These refer to coherence errors where there is a sudden topic shift in the summary.

(3) Inconsistent, Conflicting Information (\textbf{InconE}). These refer to text spans that contradict previous content.

(4) Repetition (\textbf{RepE}). These refer to text spans where content is repeated.

We show examples of these types of errors in Tab.~\ref{tab:coherence-example}.
We randomly sampled 30 examples from the test set of \texttt{GovReport}, and counted the number of text spans containing the coherence errors in the summaries generated by \textbf{\model} and \textbf{LED}. 
All examples are annotated by three of the authors.\footnote{The Krippendorff’s alpha~\citep{Krippendorff2011ComputingKA} is 0.5719.}
We anonymized the examples for a fair comparison.
The results are shown in Tab.~\ref{tab:human}.
Aligned with the findings in \citet{Goyal2022SNaCCE}, we found that both LED and \model make a non-trivial amount of errors.
However, \model is able to make fewer errors for each of the error types except for the InconE error type.

\begin{table}[t!]
\centering
\small
\begin{tabular}{@{\extracolsep{-1pt}}lccccc}
\toprule
\textbf{System} & \textbf{RefE} & \textbf{TopicE} & \textbf{InconE} & \textbf{RepE} & \textbf{Total} \\
\midrule
LED &  41.7 &  19 &  \textbf{8.3}   &  9.7 & 78.7 \\
\model & \textbf{32.3}  & \textbf{14}  &  10  & \textbf{8.3} & \textbf{64.7} \\
\bottomrule
\end{tabular}
\vspace{-2mm}
\caption{Human Evaluation for Coherence on \texttt{GovReport}. We report the number of different coherence errors made by \model and LED on 30 examples (averaged across three annotators). \textbf{RefE}: Missing Information/Reference about an Event/Object. \textbf{TopicE}:  Abrupt Transition from the Previous Topic. \textbf{InconE}: Inconsistent, Conflicting Information. \textbf{RepE}: Repetition.}
\label{tab:human} 
\vspace{-3mm}
\end{table}

\subsection{Case Study: Long-Distance Dependencies}
\label{subsec:case}

A global context can be much more important in the presence of long-distance dependencies for text summarization models~\citep{fernandes2018structured, xu-etal-2020-discourse}.
To study this phenomenon, we leverage the notion of \textit{sentence fusion}~\citep{barzilay-mckeown-2005-sentence} to investigate sentence-level dependencies.
Specifically, following \citet{lebanoff-etal-2019-analyzing, lebanoff-etal-2019-scoring}, we define a \textbf{fusion sentence} in the \textit{reference summary} to be a sentence that has significant overlaps with two or more sentences\footnote{We focus on the case of two sentences.} in the \textit{source document}.
Then, we define two sentences $\hat{s}_1, \hat{s}_2$ in the source document $D$ to be \textbf{interdependent} if they have the most significant contribution to a fusion sentence $h$:
\begin{equation}
\small
    (\hat{s}_1, \hat{s}_2) := \argmax_{(s_i, s_j), s_i, s_j \in D} \mathrm{ROUGE}_{\mathrm{Recall}} (h, s_i \oplus s_j).
\end{equation}
More details can be found in Appendix~\ref{sec:appendixa}.

\begin{figure}[t!]
    \centering
    \includegraphics[width=0.9\linewidth]{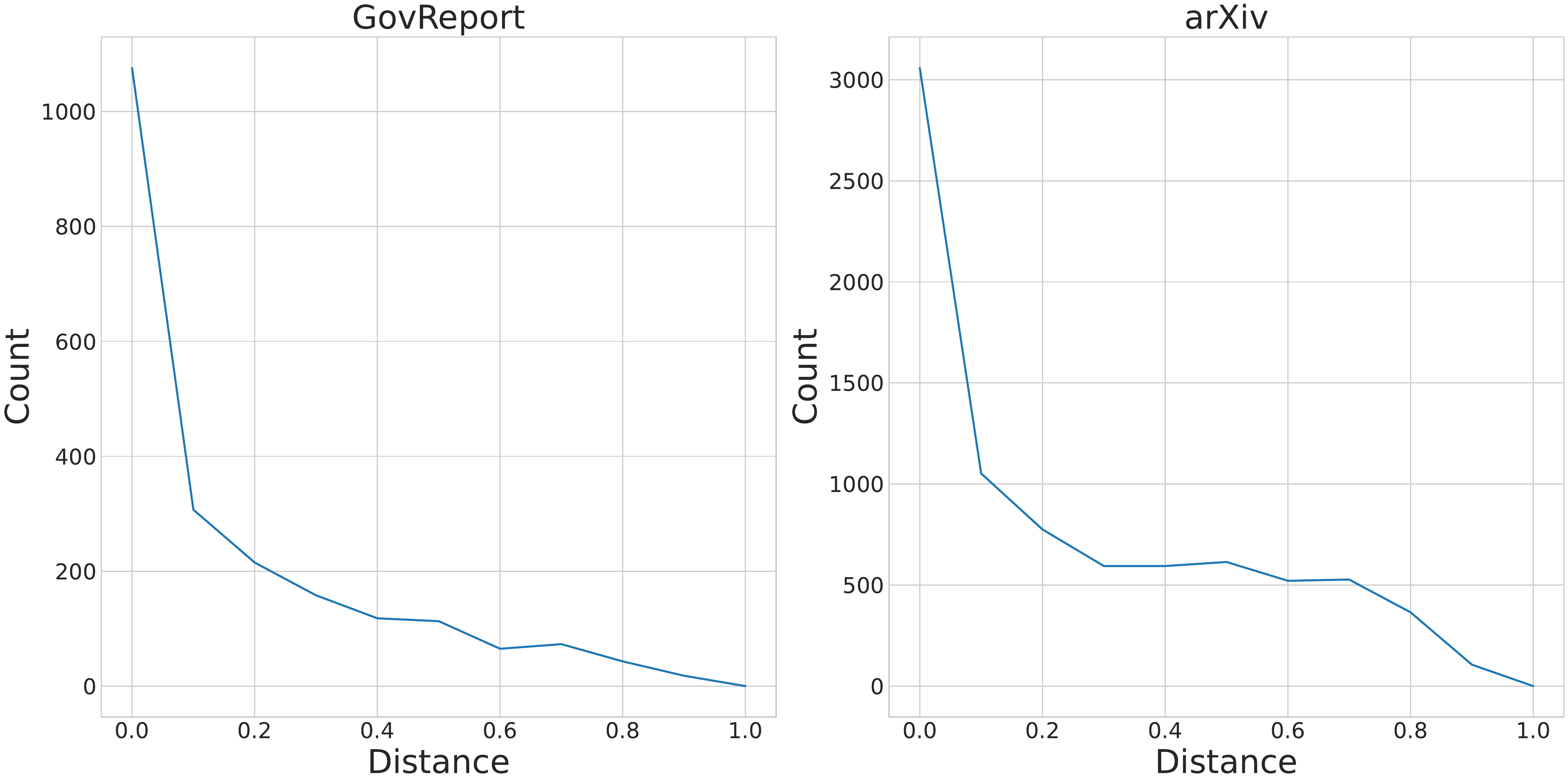}
    \caption{Number of interdependent sentences with different distances on \texttt{GovReport} and \texttt{arXiv} datasets. X-axis represents the ratio of sentence distances normalized by the number of sentences in the document.}
    \label{fig:long}
\vspace{-5mm}
\end{figure}

We found that \model can fail to capture the dependencies where two interdependent sentences are far away from each other. 
We show such an example in Tab.~\ref{tab:example}, where the 14th sentence and 410th sentence in the source document both contribute to the same fusion sentence.
\model's output only captures the information in the 14th sentence.
However, the impact of the potential failures is restricted.
As shown in Fig.~\ref{fig:long}, there are much fewer interdependent sentence pairs with long distances.

\begin{table}[t!]
    \scriptsize
    \centering
    \extrarowheight=\aboverulesep
    \addtolength{\extrarowheight}{\belowrulesep}
    \aboverulesep=0pt
    \belowrulesep=0pt
    \begin{tabular}{@{} p{0.2\linewidth} p{0.7\linewidth}}
     \toprule
\textbf{Fusion \newline Sentence} & \cellcolor{gray!25} ED issued a notice of \textit{proposed rulemaking} in \textbf{late 2018}, after revoking some of its previous guidance to schools in 2017. \\
\midrule
 \textbf{14th Source Sentence} &  \cellcolor{gray!10} And ED recently issued another notice of \textit{proposed rulemaking}, after having revoked some of its prior guidance to schools in 2017. \\
\midrule
 \textbf{410th Source Sentence} &  \cellcolor{gray!25}  On \textbf{November 29, 2018}, ED issued a notice of \textit{proposed rulemaking} in the Federal Register. \\
\midrule
  \textbf{\model \newline Output} & \cellcolor{gray!10} ED recently issued another notice of proposed rulemaking, after having revoked some of its prior guidance to schools in 2017.\\
  \bottomrule
\end{tabular}
\caption{Case Study on \texttt{GovReport} about long-distance dependencies. Both 14th and 410th sentences contribute to the same reference sentence. \model's output fails to capture this long-distance dependency.}
\label{tab:example}
\vspace{-2mm}
\end{table}

\section{Conclusions}

We empirically investigate three kinds of locality in abstractive text summarization by using them as important inductive biases.
Using a new abstraction of viewing the input document as a series of \textit{pages}, our model emphasizes the role of locality in both encoding and decoding stages.
The experimental results show that our model has strong performance by following the principle of locality. 
We also show that it is important to select the suitable kind of locality for different application scenarios.
We note that the fact that our model has better or competitive performance comparing with the models equipped with efficient attention modules suggests that those models may fall short of their designing objectives.
Therefore, for future work, our findings call for more rigorous examinations of the memory-efficient abstractive summarization models that aim to capture global features (e.g. long-distance dependencies) and maintain a global input context.

\section{Limitations}

\textbf{Computation Resources} While our approach can reduce the memory footprint of full-attention models, it still requires GPUs with large memory sizes (e.g. 48 GBs) and long time (more than 7 days with a single GPU)
to train our model.
We note that our model has a similar memory footprint as the efficient-attention models such as Longformer~\citep{DBLP:journals/corr/abs-2004-05150}.
Therefore, the requirement of computation resources is a common challenge in long text summarization.

\textbf{Long-Distance Dependencies} The inductive bias of our approach is to emphasize the role of \textit{locality} in abstractive text summarization. 
As a result, our approach can fail to capture long-distance dependencies. 
We have discussed this potential problem in \S\ref{subsec:case}.
While we have shown that the ratio of sentence-level long-distance dependencies are relatively low in the datasets we investigated for this work, it is worthwhile to be aware of this limitation when extending our method to other datasets.

\textbf{Human Evaluation} While we have presented a fine-grained human evaluation on summary coherence in \S\ref{subsec:human}, there are other important aspects of summary quality such as factual consistency~\citep{maynez-etal-2020-faithfulness}.
However, it is even a more non-trivial task to evaluate an input-document-based aspect such as factual consistency on the datasets we used as it requires reading the entire input documents which can be more than 10K words long and having domain-specific knowledge to understand the context of scientific papers or government reports.
We believe the research of long text summarization will benefit greatly from better human and automatic evaluation.

\section*{Acknowledgements}
We thank the anonymous reviewers for helpful suggestions.
This work is supported in part
by a grant from Microsoft Research.

\bibliography{anthology,custom}
\bibliographystyle{acl_natbib}

\appendix

\section{Experimental Settings}
\label{sec:appendix-setting}

\begin{table}[h]
  \centering
  \small
    \begin{tabular}{@{\extracolsep{0.5pt}}lccccc}
    \toprule
    \multirow{2}{*}{Datasets} & \multicolumn{3}{c}{\# Examples} & \multicolumn{2}{c}{Avg. Tokens} \\
    \cmidrule{2-4} \cmidrule{5-6}
    & Train & Valid & Test & Doc. & Sum. \\
    \midrule
    arXiv & 203K & 6.4K & 6.4K & 8154.3 & 197.8 \\
    PubMed & 120K & 6.7K & 6.6K & 3983.6 & 261.3 \\
    GovReport & 17.5K & 973 & 974 & 10726.1 & 681.6\\
    MultiNews & 45.0K & 5.6K & 5.6K & 2526.4 & 277.2\\
    \bottomrule
    \end{tabular}%
  \caption{Dataset Statistics. We report the average number of tokens generated by the BPE tokenizer~\cite{sennrich-etal-2016-neural} used by BART~\cite{lewis-etal-2020-bart} on the \textit{validation} set. For MultiNews dataset, we report the \textit{sum} of lengths of the individual source document in a document cluster as it is a multi-document dataset.}
  \label{tab:data}%
\end{table}%

\subsection{Datasets Statistics}

We report the dataset statistics in Tab.~\ref{tab:data}.

\subsection{Implementation Details}

We use the Adam optimizer~\citep{DBLP:journals/corr/KingmaB14} with learning rate scheduling as follows:
\begin{equation}
\small
    lr = 2 \times 10^{-3} \min(\textrm{step}^{-0.5}, \textrm{step}\cdot\textrm{warmup}^{-1.5}).
\end{equation}
$\textrm{warmup}$ is the number of warmup steps, which is set to 10000.
$\textrm{step}$ is the number of update steps taken so far.
Our models are trained on one NVIDIA A6000 GPU, and it takes around 5-25 hours (depending on the size of the dataset) for one training epoch.
All models converged in 10 epochs.
For ROUGE score computation, we use the summary-level ROUGE-L score which is the default choice of the standard ROUGE Perl script.

\section{Long-Distance Dependencies}
\label{sec:appendixa}

We define two sentences $\hat{s}_1, \hat{s}_2$ in the source document $D$ to be \textbf{interdependent} if they have the most significant contribution to a fusion sentence $h$ in the reference summary:
\begin{equation}
\small
    (\hat{s}_1, \hat{s}_2) := \argmax_{(s_i, s_j), s_i, s_j \in D} \mathrm{ROUGE}_{\mathrm{Recall}} (h, s_i \oplus s_j).
\end{equation}
where we use ROUGE Recall to measure the sentence contribution by viewing $h$ as the reference.
We define two filtering rules:
\begin{equation}
\label{eq:rule_1}
\small
    \mathrm{ROUGE}(h, s) > t_1,
\end{equation}
\vspace{-25pt}
\begin{equation}
\label{eq:rule_2}
\small
    \mathrm{ROUGE} (h, \hat{s}_1 \oplus \hat{s}_2) - \mathrm{ROUGE}(h, s) > t_2,
\end{equation}
where $s \in \{\hat{s}_1, \hat{s}_2\}$. 
$t_1$ and $t_2$ are two threshold values which are set to 20 and 10 respectively based on our empirical observations.  
Eq.~\ref{eq:rule_1} ensures that each sentence has a non-trivial overlap with the fusion sentence, while Eq.~\ref{eq:rule_2} ensures that each sentence has a unique contribution.

\end{document}